%% file: main.tex
\documentclass[conference]{IEEEtran}
\IEEEoverridecommandlockouts
\usepackage{cite}
\usepackage{amsmath,amssymb,amsfonts}
\usepackage{graphicx}
\usepackage{textcomp}
\usepackage{algorithm}
\usepackage{algpseudocode}
\usepackage{hyperref}
\usepackage{footmisc}
\usepackage{multirow}
\usepackage{url}
\usepackage{xcolor}
\newtheorem{definition}{Definition}
\def\BibTeX{{\rm B\kern-.05em{\sc i\kern-.025em b}\kern-.08em
    T\kern-.1667em\lower.7ex\hbox{E}\kern-.125emX}}
\begin{document}

\title{Representation Learning for Spatial Graphs}

\author{
    \IEEEauthorblockN{Zheng Wang\IEEEauthorrefmark{1}, Ce Ju\IEEEauthorrefmark{2}, Gao Cong\IEEEauthorrefmark{1}, Cheng Long\IEEEauthorrefmark{1}}
    \IEEEauthorblockA{\IEEEauthorrefmark{1}School of Computer Science and Engineering, Nanyang Technological University, Singapore\\
                      \IEEEauthorrefmark{2}IDG, Baidu Inc., Beijing, China\\
    \{wang\_zheng, gaocong, c.long\}@ntu.edu.sg, juce@baidu.com}
}

\maketitle
\begin{abstract}
\input{abstract}

\end{abstract}

\input{introduction}
\input{related}

\input{problem}
\input{method}

\input{experiments}

\input{conclusion}

\bibliography{ref}
\bibliographystyle{IEEEtran}

\end{document}

%% file: abstract.tex
Recently, the topic of graph representation learning has received plenty of attention. Existing approaches usually focus on structural properties only and thus they are not sufficient for those spatial graphs where the nodes are associated with some spatial information.
In this paper, we present the first deep learning approach called s2vec for learning spatial graph representations,
which is based on denoising autoencoders framework (DAF).
We evaluate the learned representations on real datasets and the results verified the effectiveness of s2vec when used for spatial clustering.

%% file: introduction.tex
\section{introduction}
Spatial graphs are graphs where each node is associated with a location~\cite{dale2010graphs}.
One well-known instance of spatial graph is the location-based social networks (LBSN)~\cite{shi2014density,fortunato2010community,fang2018spatial},
where the users in social networks, which correspond to nodes, are associated with some location information,
e.g., online "check-ins" in Brightkite, geo-tagged tweets in Twitter, geo-tagged photo albums in Foursquare.
Some other instances of spatial graph are commonly used in ecology and evolution~\cite{dale2010graphs}.

We consider the problem of spatial subgraph embedding based on deep representation learning. Once the embedding vectors of the graphs are computed, they could be used for a variety of downstream graph analysis tasks, including spatial clustering~\cite{shi2014density}, outlier detection~\cite{fang2018spatial} and spatial classification. For instance, in social network analysis, we often gained subgraph-structured communities by performing community search or detection algorithms. Clustering these communities are very beneficial for some practical applications such as social marketing, as studied in~\cite{manchanda2015social}, people with close social relationships tend to purchase in places that are also physically close. Advertisers can target limited resources to the communities with similar spatial and structural characteristics to boost sales figures and achieve maximize revenue.

Existing graph representation learning methods can be categorized into node representation~\cite{perozzi2014deepwalk,grover2016node2vec,tang2015line,li2017deepcas} and subgraph representations~\cite{niepert2016learning,narayanan2016subgraph2vec,yanardag2015deep,Taheri2018LearningGR}.
However, almost all these prior works focus on structural analysis and do not consider the spatial features.
Some alternative approaches for spatial graph similarity search include
graph kernels~\cite{tan2014effect} and graph matching~\cite{aanjaneya2012metric},
which, however cannot scale-up for large graphs due to their high computational cost.
Additionally, these approaches do not consider the global structure of the graphs and are sensitive to noise.

In this paper, we propose an unsupervised \textbf{\underline{s}}patial graph to \textbf{\underline{vec}}tor approach called \textbf{s2vec} for learning representations of spatial subgraphs based on the LSTM denoising autoencoders framework (DAF).
Specifically, we first sample a set of paths on each subgraph via random walks.
Inspired by~\cite{li2018deep}, a spatial information aware loss function is proposed based on negative log likelihood, which captures the similarity based on spatial proximity.
Finally, we demonstrate the effectiveness of our approach by comparing our learned representations with those based some baselines when used for a spatial clustering task.
To the best of our knowledge, s2vec is the first method that supports spatial subgraph representation, making it widely applicable to various downstream graph analysis tasks.

%% file: related.tex
\section{related work}

Our work is closely related to the literature of representation learning for graphs. The traditional methods in this field represent a graph as an associated matrix (e.g., adjacency matrix) or a collection of nodes and edges.
In recent years, inspired by the success of word2vec~\cite{mikolov2013efficient}, modern learning methods attempt to embed nodes into high-dimensional vectors in a continuous space so that nodes with similar representation vectors share similar structural properties such as DeepWalk~\cite{perozzi2014deepwalk}, node2vec~\cite{grover2016node2vec}, LINE~\cite{tang2015line} and DeepCas~\cite{li2017deepcas}.

Another line of related work comes from the representation of subgraph structures~\cite{niepert2016learning,narayanan2016subgraph2vec,yanardag2015deep,Taheri2018LearningGR}.
Many of the approaches are inspired by the huge success of representation learning and deep neural networks applied to various domains. For example, subgraph2vec~\cite{narayanan2016subgraph2vec} borrows the key idea from document embedding methods. The biggest difference between our method and the above works is that these
model structures are designed without considering the spatial information and can not be directly used for spatial graph embedding.

In addition, some new graph kernels such as Weisfeiler-Lehman subtree kernel (WL)~\cite{tan2014effect} and Deep Graph Kernel (DGK)~\cite{yanardag2015deep} have been proposed to characterize the similarity of different network structures. Others are motivated by representation learning of images to embed graph using convolutional neural networks (CNNs)~\cite{zhang2018end,niepert2016learning}.


%% file: problem.tex
\section{DEFINITIONS AND PRELIMINARIES}
In this section, we present definitions and preliminaries necessary to understand the problem solved and the denoising autoencoders model used in our solution.

\subsection{Definitions}

\begin{definition}[Spatial Graph]
\label{def:Geo Social Graph}
Let $G=(V, E)$ be a graph,
where $V$ denotes its vertex set and $E\subseteq (V \times V)$ denotes its edge set. $G$ is called a spatial graph if
each vertex $v \in V$ corresponds to a tuple $(id, x, y)$, where $id$ is the identification number of $v$ and $(x,y)$ is the location of $v$.
\end{definition}

%

\begin{definition}[Gromov-Hausdorff Distance]
\label{def:Gromov-Hausdorff Distance}
Given two spatial graphs $G_1=(V_1, E_1) $ and $G_2 = (V_2, E_2)$, the one-sided graph distance from $G_1$ to $G_2$ is defined as:
\[ \tilde{d}_H(G_1, G_2) := \max_{v_1\in V_1} \min_{v_2 \in V_2} d(v_1, v_2),
\]
where $d(\cdot, \cdot)$ is the distance on Euclidean space. The bidirectional graph distance between $G_1$ and $G_2$ is defined as
\[
d_H(G_1, G_2) := \max (\tilde{d}_H(G_1, G_2), \tilde{d}_H(G_2, G_1))
\]
is then called Gromov-Hausdorff distance of spatial graph.
\end{definition}

%

\subsection{Problem statement}
Given a set of spatial subgraphs $\mathbb{G}=\{G_{1}, G_{2},...\}$,
we aim to compute an embedding vector of each subgraph such that
if two subgraphs are spatial similar w.r.t. Gromov-Hausdorff distance,
then they are also similar w.r.t. the Euclidean distance based on their embedding vectors.
Specifically, given small enough $\delta$, $\epsilon > 0$ and for any spatial subgraph $G_i$ and $G_j$,
\[
\text{if } d_H(G_i, G_j) \leq \delta, \text{ we have } d(\phi(G_i), \phi(G_j)) \leq \epsilon,
\]
where $d_H(\cdot, \cdot)$ is Gromov-Hausdorff distance, $d(\cdot, \cdot)$ is Euclidean distance and function $\phi(\cdot)$ is our embedding algorithm. Suppose $n$ is the embedding dimension, the matrix of representations of the subgraph sets $\mathbb{G}$ is denoted as $\Phi\in\mathbb{R}^{|\mathbb{G}| \times n}$ in the following sections. Note that we did not preserve the structures of these subgraphs and assume they are similar. For example, they are all the same $k$-core subgraphs obtained by community search algorithms,

\subsection{Preliminaries of denoising autoencoders framework}
\begin{figure}
  \centering
  \includegraphics[width=5cm]{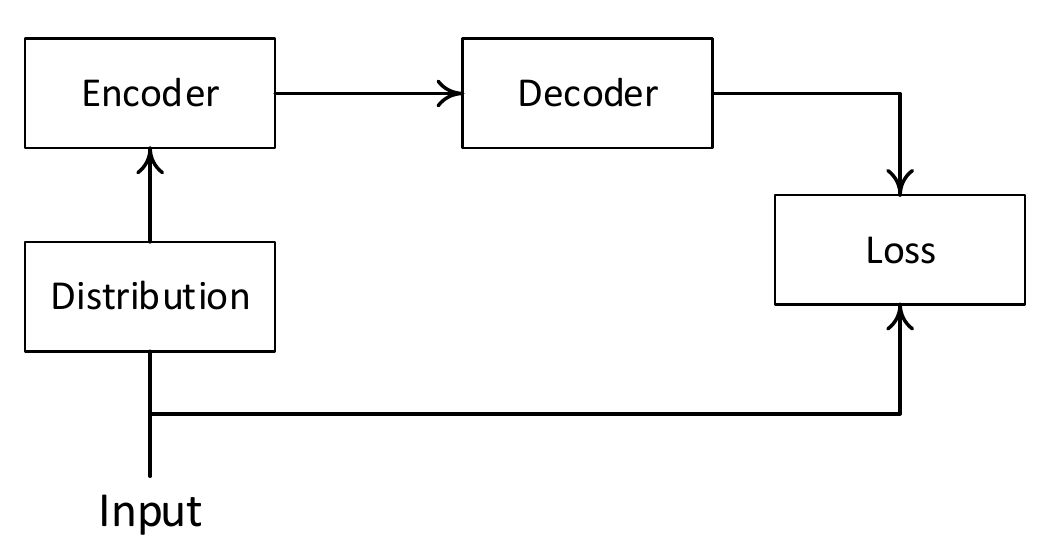}\\
  \caption{Denoising Autoencoders Model}\label{dae}
\end{figure}

We briefly introduce the denoising autoencoders framework, which is proposed by Vincent et al.~\cite{vincent2008extracting}. The principle behind it is to be able to reconstruct data from an input of corrupted or noisy data. In order to force the hidden layer to discover more robust features and prevent it from simply learning the identity, and thus the output will be a more refined version of the input data. Precisely, the denoising autoencoders framework has the following steps, as shown in Figure~\ref{dae}.
\begin{itemize}
  \item \textbf{Distribution:} Corrupt the initial input $x$ by sampling from a conditional distribution $q_{\mathcal{D}}(\tilde{x}|x)$ which stochastically maps input data $x$ to a corrupted version $\tilde{x}$.
  \item \textbf{Encoder:} Map the corrupted input $\tilde{x}$ to a latent representation $h: =f_{\theta}(\tilde{x})$.
  \item \textbf{Decoder:} Reconstruct a data $z := g_{\theta^*}(h)$ from the latent representation $h$.
  \item \textbf{Loss:} Compute the average reconstruction error $\mathcal{L}(x,z)$ over a training set.
\end{itemize}

The objective function of denoising autoencoders framework is
\[
\arg \min_{\theta^*,\theta} \mathbb{E}_{q^0(x, \tilde{x})} \mathcal{L}(x, z)
\]
where $z = g_{\theta^*} \big(f_{\theta}(\tilde{x})\big)$ and $q^0(x, \tilde{x})$ denotes the empirical joint distribution associated to our training inputs. Minimizing the objective function by stochastic gradient descent method, the denoising autoencoders framework reconstruct the uncorrupted data from the corrupted one.

%

%

%% file: method.tex
\section{METHOD}
Our proposed method for solving this problem is inspired by the denoising autoencoders framework and techniques of Seq2Seq and the spatial proximity computation.
\subsection{Random walk sampling}
In this subsection, we will introduce our path sampling method by which we sample a group of paths on each subgraph $G \in \mathbb{G}$. The path sampling method is simplified from a popular graph sampling method called Random Walk with Jump~\cite{li2017deepcas}. The reason for the simplification of our sampling is in order to sample a set of connected paths. We walk from a starting node $v_{t:=start}$ picked uniformly from the whole subgraph. In the next step, we have a given probability $\mathbb{P}$ to walk to one of the adjacent node selected uniformly at random. The uniform selection means in the following way that supposes the degree of the current node $v$ is $deg(v)$, then we have the probability transition from the current node $v$ to its adjacent node $u$ is
\[
p(u\in N(v)|v)=\frac{deg(u)+\eta}{\Sigma_{s \in N(v)}(deg(s)+\eta)},
\]
where $N(v)$ denotes the set of $v'$s outgoing neighbors. $\eta$ is a smoother. Otherwise, we need to stay at the current node with probability $1-\mathbb{P}$. The walk stops at the end node $v_{t:=end}$ after enough number of nodes are visited. According to our construction of the path, each path of the subgraph could be represented as $\mathcal{T}_G:= v_{t:=start} \cdots v_{t:=end}$,
which is also regarded as a user visiting order-spatial sequence. In the following sections, for each subgraph $G \in \mathbb{G}$, $set(\mathcal{T}_G)$ corresponds to the set containing all sampling paths on the subgraph $G$.

\subsection{The latent representation of subgraph}
For each subgraph $G\in\mathbb{G}$, we construct a corrupted subgraph $\tilde{G}$ which is added random noise $\delta \cdot \mathcal{N}(0, 1)$ on each node of the initial subgraph $G \in \mathbb{G}$, where $\delta > 0$ is the magnitude of noise and $\mathcal{N}(0, 1)$ is the standard normal distribution. Thus, the Gromov-Hausdorff distance between the initial subgraph $G$ and its corresponding corrupted subgraph $\tilde{G}$ is bounded by the magnitude $\delta$.
This can be proved directly by definition. Also, we construct the corrupted path set $set(\mathcal{T}_{\tilde{G}})$ of $\tilde{G}$. Each path $\mathcal{T}_{\tilde{G}}$ of $set(\mathcal{T}_{\tilde{G}})$ only differs the Gaussian noisy displacement on each node of the initial path $\mathcal{T}_G$ of $G$.

In the sequence encoder-decoder model, we need to input the discrete token into the model and thus we partition the spatial space into cells of equal size. For each node on the path, if it falls into one of the cells and then we provide this node a token.
We only keep the cells which are sufficiently hit by sample points. These cells are referred to as hot cells $C$.
The sequence of token of the path $\mathcal{T}_G$ is denoted as $\mathcal{S}_G = \{s_1,s_2,\cdots\}$. The sequence set is denoted as $set(\mathcal{S}_G)$.

According to the denoising autoencoders framework, we pick LSTM as the encoding function $f_{\theta}(\cdot):=LSTM^{enc}_{\theta}(\cdot)$. Then, the time-varying hidden vector $h^{enc}_t$ is
\[
h^{enc}_t :=LSTM^{enc}_{\theta}(\tilde{s}_{t}, h^{enc}_{t-1}),
\]

where $\tilde{s}_{t}$ means the corrupted version at timestamp $t$. We let another LSTM as the decoding function
$g_{\theta^*}(\cdot):=LSTM^{dec}_{\theta^*}(\cdot)$, then 
\[
h^{dec}_t :=LSTM^{dec}_{\theta^*}(s_{t-1}, h^{dec}_{t-1}),
\]
the objective function of our algorithm is inspired by the classic negative log likelihood loss~\cite{platt1999probabilistic},
\[
\arg\min_{\theta,\theta^*}  \mathbb{E}_{q^0(\mathcal{S}_G, \mathcal{S}_{\tilde{G}})} \mathcal{L}\Big(\mathcal{S}_G,  g_{\theta^*}(f_{\theta}(\mathcal{S}_{\tilde{G}}\big)\Big).
\]
The reconstruction error in our loss function is specifically defined as the spatial proximity aware loss~\cite{li2018deep},
\[
 \mathcal{L}:= -\sum_{t=1}^{|S_G|} \sum_{u \in C} w_{us_t} \log{\frac{\exp{W_u^T\cdot h^{dec}_t}}{\sum_{v\in C}\exp{(W_v^T\cdot h^{dec}_t)}}},
\]
where the coefficients of the polynomial
\[
  w_{us_t} = \frac{\exp{(-||u-s_t||_2)}}{\sum_{v\in C} \exp{(-||v-s_t||_2)}}
\]
is the spatial proximity weight on cell $v\in C$ when decoding target $s_t$. $W_u^T$ is the projection matrix for cell $u$ that projects $h_t$ from the hidden state spaces into the token space. And, $||u-s_t||_2$ denotes the Euclidean distance between the centroid coordinates of the cells.

Eventually, the graph embedding function $\phi(G)$ is defined as the average of the latent representation of the sampling path for each subgraph $G$,
\[
\phi(G) := \frac{1}{|set(\mathcal{S}_G)|}\sum_{\mathcal{S}_G\in set(\mathcal{S}_G)} f_{\theta}(\mathcal{S}_G).
\]

Thus, our desired subgraph matrix representation is
\[
\Phi := \Big(\phi(G_1), \cdots, \phi(G_{|\mathbb{G}|})\Big) \in \mathbb{R}^{|\mathbb{G}|\times n},
\]
where $n$ is the embedding dimension.

\subsection{Algorithm overview}
\label{overview}
Algorithm~\ref{alg:framework} presents the framework. The input includes a set of spatial subgraphs $\mathbb{G}=\{G_{1}, G_{2},...\}$, the embedding size $n$, learning model $M_{\theta}$ with encoding function $f_{\theta}$ and decoding function $g_{\theta}$, where $\theta$ is the global parameters and learning rate $\alpha$ (line 1-4). During the iterative training process (line 5-11), we first sample a set of paths of vertices $set(\mathcal{T}_G)$ from the set of subgraphs (line 6). Similar to DeepWalk~\cite{perozzi2014deepwalk}, the sampling process could be generalized as performing a random walk. Then, we stochastically map the original input $set(\mathcal{T}_G)$ to a noisy version $set(\mathcal{T}_{\tilde{G}})$ by adding a Gaussian noise which is subject to the conditional distribution $q_{D}$ (line 7) and perform the tokenization to get $set(\mathcal{S}_{G})$ and $set(\mathcal{S}_{\tilde{G}})$ respectively (line 8). Next, we get the reconstruction $\mathcal{S}_{\hat{G}}$, which is computed by decoder component of the model $\mathcal{M}_{\theta}$ and an optimizer such as stochastic gradient descent (SGD) is used to optimize the parameters $\theta$ (line 9-10). Finally, a vector representation of each subgraph is computed via the learned model $\mathcal{M}_{\theta}$ (line 12-15).
Note that the ultimate output $\Phi$ is a matrix of vector representations of spatial subgraphs.

\begin{algorithm}[]
\caption{Algorithm framework}
\footnotesize{
\label{alg:framework}
\begin{algorithmic}[1]
        \Require
        \State $\mathbb{G}=\{G_{1}, G_{2},...\}$: a set of spatial graphs need to be embedded
        \State $n$: the embedding size
        \State Learning model $\mathcal{M}_{\theta}$ with encoder $f_{\theta}$ and decoder $g_{\theta}$
        \State $\alpha$: learning rate
        \Repeat
        \State sample $set(\mathcal{T}_G)$ by random walking from the set of subgraphs
        \State get noisy version $ set(\mathcal{T}_{\tilde{G}}) \sim q_{D}( set(\mathcal{T}_{G})| set(\mathcal{T}_{\tilde{G}}) )$
        \State get $set(\mathcal{S}_{G})$ and $set(\mathcal{S}_{\tilde{G}})$ from $set(\mathcal{T}_{G})$ and $set(\mathcal{T}_{\tilde{G}})$
        \State compute reconstruction $ \mathcal{S}_{\hat{G}} = g_{\theta}(f_{\theta}(\mathcal{S}_{\tilde{G}}\big)$ from $\mathcal{M}_{\theta}$
        \State $\theta \leftarrow \theta - \alpha\nabla_{\theta}\mathcal{L}(\mathcal{S}_{G}, \mathcal{S}_{\hat{G}})$
        \Until{No improvement on validation set}

        \For{each $G_{i} \in \mathbb{G}$}
            \State get representation $\phi(G_{i})$ from $\mathcal{M}_{\theta}$
            \State $\Phi \leftarrow \phi(G_{i})$
        \EndFor
        \Ensure
        Matrix of vector representations of spatial graphs $\Phi\in\mathbb{R}^{|\mathbb{G}| \times n}$
\end{algorithmic}}
\end{algorithm}

%% file: experiments.tex
\section{EXPERIMENTS}
In this section, we use our representations for the task of spatial graph clustering on four real datasets and show the effectiveness of our model s2vec compared against several baseline approaches.

\subsection{Datasets}
The experiments are conducted on four real datasets:
Brightkite\footnote{http://snap.stanford.edu/data/index.html\label{fn:repeat}},
Gowalla\textsuperscript{\ref{fn:repeat}}, Flickr\footnote{https://www.flickr.com/}
and Foursquare\footnote{https://archive.org/details/201309\_foursquare\_dataset\_umn}.
For all the datasets, each vertex represents a user and each link represents the friendship between two users. The statistics of each dataset are shown in Table~\ref{tab:dataset}, where the average degree $\widehat d$, the average betweenness centrality \textbf{BC} and the clustering coefficient \textbf{CC} of each vertex in the dataset are included. We randomly select 200 subgraphs of each dataset
among those found by a community search algorithm~\cite{fang2018spatial}.

\begin{table}[ht]
  \centering
  \scriptsize
  \caption {Dataset statistics.} \label{tab:dataset}
  \vspace{-0.2cm}
  \begin{tabular}{c|c|r|r|c|c|c}
     \hline
          {\bf Type}     & \multicolumn{1}{c|}{\textbf{Name}}
                         & \multicolumn{1}{c|}{\textbf{Vertices}}
                         & \multicolumn{1}{c|}{\textbf{Edges}}
                         & \multicolumn{1}{c|}{\textbf{\emph{{$\widehat d$}}}}
                         & \multicolumn{1}{c|}{\textbf{BC}}
                         & \textbf{CC}\\
     \hline\hline
          \multirow{4}*{Real} & Brightkite     &51,406    &197,167   & 7.67 &7.34E-5 &0.1795\\
     \cline{2-7}
          &Gowalla        &107,092   &456,830   & 8.53 &3.40E-5 &0.2487\\
     \cline{2-7}
          &Flickr         &214,698   &2,096,306 & 19.5 &1.65E-5 &0.1113\\
     \cline{2-7}
          &Foursquare     &2,127,093 &8,640,352 & 8.12 &1.68E-6 &0.1044\\
     \hline
  \end{tabular}
\end{table}

\subsection{Baselines}
We compare s2vec with two baselines of graph representation, the first baseline is similar to Principal Component Analysis (PCA)~\cite{wold1987principal}, we arrange the spatial features of each vertex in the subgraph according to the order of longitude and latitude and then expand into a high-dimensional vector. In this case, vectors of inconsistent length are padded by a special value such as zero. After that, we perform PCA dimensionality reduction on all subgraphs to obtain new vector representations. Note that this algorithm considers the graph as a set of spatial vertices without structural features.

The second baseline is s kind of ablation of s2vec, adapting the sequence-to-sequence learning framework for autoencoding by using the same sequence for both the input and output~\cite{li2015hierarchical,Taheri2018LearningGR}. We denote these two baselines as \textbf{PCA} and \textbf{Vanilla-s2vec (V-s2vec)} respectively.

\subsection{Spatial Graph Clustering}
Given a set of spatial graphs $\mathbb{G}$, the goal of graph clustering is to group graphs with similar structure and spatial features together. s2vec's representation vectors could be used along with conventional clustering algorithms such as DBSCAN with Euclidean distance for this purpose.

The lack of ground-truth makes it a challenging problem to evaluate the effectiveness of spatial graph clustering. To overcome it, we use the Gromov-Hausdorff distance to measure the similarity of spatial graphs and perform DBSCAN algorithm to find the results as its ground-truth.

\textbf{Evaluation Metric.} In order to quantitatively measure the clustering accuracy, a standard clustering evaluation metric, namely, Adjusted Rand Index (ARI)~\cite{rand1971objective} is used. The ARI values lie in the range $[-1, 1]$. A higher ARI means a higher correspondence to the ground-truth results.
We convert it into a percentage value for easy understanding.

The results of spatial graph clustering using s2vec and other baselines are presented in Table~\ref{tab:spacluster}. We observe that s2vec outperforms all the other compared approaches highly significantly. In particular, it outperforms the PCA techniques by at least 10\%. This is because
s2vec fully considers maintaining similarity in original space.

\begin{table}[ht]
  \centering
  \caption {Spatial Graph Clustering.} \label{tab:spacluster}
  \vspace{-0.2cm}
  \begin{tabular}{c|c|c|c|c}
     \hline
    \multicolumn{1}{c|}{Dataset}
    &\multicolumn{1}{c|}{Brightkite}
    &\multicolumn{1}{c|}{Gowalla}
    &\multicolumn{1}{c|}{Flickr}
    &Foursquare\\
     \hline
       PCA     &49.76	&49.45 &49.99 &48.85\\
     \hline
       V-s2vec &57.14	&59.81 &69.59 &54.60\\
     \hline
      s2vec    &\textbf{62.49}	&\textbf{63.07} &\textbf{73.64} &\textbf{63.61}\\
     \hline
  \end{tabular}
\end{table}

%% file: conclusion.tex
\section{conclusion and future work} 
In this paper, we learn the latent representation of the spatial subgraphs in the denoising autoencoders framework and apply our latent representations in the spatial graph clustering task achieving the effective results. The problem of spatial graph representation is the first time to be proposed and has a great potential application value. In the future, we will consider a more complex graph structure with more information on its nodes. 